\documentclass[11pt,a4paper]{article}
\usepackage[dvipsnames]{xcolor}
\usepackage[hyperref]{acl2018}
\usepackage{times}
\usepackage{latexsym}
\usepackage{url,graphicx}

\usepackage{CJK,algorithm,algorithmic,amssymb,amsmath,array,epsfig,graphics,multirow,array,float,subfigure,verbatim,epstopdf}
\usepackage{enumitem}
\usepackage{color,soul}
\usepackage{hhline}
\usepackage{multirow}
\usepackage{xcolor}
\usepackage{hyperref}
\newcommand{\RNum}[1]{\uppercase\expandafter{\romannumeral #1\relax}}
\aclfinalcopy 
\def\aclpaperid{217} 

\usepackage{xparse}
\NewDocumentCommand{\heng}{ mO{} }{\textcolor{OrangeRed}{\textsuperscript{\textit{Heng}}\textsf{\textbf{\small[#1]}}}}
\NewDocumentCommand{\lifu}{ mO{} }{\textcolor{pink}{\textsuperscript{\textit{Lifu}}\textsf{\textbf{\small[#1]}}}}
\NewDocumentCommand{\zhihao}{ mO{} }{\textcolor{brown}{\textsuperscript{\textit{Zhihao}}\textsf{\textbf{\small[#1]}}}}
\NewDocumentCommand{\spencer}{ mO{} }{\textcolor{blue}{\textsuperscript{\textit{Spencer}}\textsf{\textbf{\small[#1]}}}}
\NewDocumentCommand{\qingyun}{ mO{} }{\textcolor{Purple}{\textsuperscript{\textit{Qingyun}}\textsf{\textbf{\small[#1]}}}}
\NewDocumentCommand{\boliang}{ mO{} }{\textcolor{green}{\textsuperscript{\textit{Boliang}}\textsf{\textbf{\small[#1]}}}}

\title{Paper Abstract Writing through Editing Mechanism}
\author{
Qingyun Wang$^{1*}$, \ Zhihao Zhou$^{1}$\thanks{$^*$Qingyun Wang and Zhihao Zhou contributed equally to this work.}, \ Lifu Huang$^1$,  \textbf{Spencer Whitehead}$^1$, \\ \textbf{Boliang Zhang}$^1$, \ \textbf{Heng Ji}$^1$, \ \textbf{Kevin Knight}$^2$ \\
$^{1}$ Rensselaer Polytechnic Institute \\
{\tt jih@rpi.edu} \\
$^{2}$ University of Southern California \\
{\tt knight@isi.edu}
}

\date{}

\begin{document}
\maketitle

\begin{abstract}

We present a paper abstract writing system based on an attentive neural sequence-to-sequence model that can take a title as input and automatically generate an abstract. We design a novel Writing-editing Network that can attend to both the title and the previously generated abstract drafts and then iteratively revise and polish the abstract. With two series of Turing tests, where the human judges are asked to distinguish the system-generated abstracts from human-written ones,
our system passes Turing tests by junior domain experts at a rate up to 30\% and by non-expert at a rate up to 80\%.\footnote{The datasets and programs are publicly available for research purpose \url{https://github.com/EagleW/Writing-editing-Network}}

\end{abstract}
\section{Introduction}


\emph{Routine writing}, such as writing scientific papers or patents, is a very common exercise.
It can be traced back to the ``\textit{Eight legged essay}'', an austere writing style in the Ming-Qing dynasty.\footnote{https://en.wikipedia.org/wiki/Eight-legged\_essay} 
We explore automated routine writing, with
paper abstract writing as a case study. Given a title, we aim to automatically generate a paper abstract. 
We hope our approach can serve as an assistive technology for human to write paper abstracts more efficiently and professionally, by generating an initial draft for human’s further editing, correction and enrichment.

A scientific paper abstract should always \textbf{focus on the topics} specified in the title.
However, a typical recurrent neural network (RNN) based approach easily loses focus. Given the title ``\textit{An effective method of using \textbf{Web} based \textbf{information} for \textbf{Relation Extraction}}'' from~\newcite{Keong2008}, we compare the human written abstract and system generated abstracts in Table~\ref{table:abstracts}. The \textit{LSTM\_LM} baseline generated abstract misses the key term ``\textit{Web}'' mentioned in the paper title. We introduce a title attention~ \citep{atten14, luong2015effective} into a sequence-to-sequence model~\cite{seq2seq14,cho2014learning} to guide the generation process so the abstract is topically relevant to the given title, as shown in the ``Seq2seq with attention'' row of Table~\ref{table:abstracts}.

Previous work usually models natural language generation as a one-way decision problem, where models generate a sequence of tokens as output and then moves on, never coming back to modify or improve the output. 
However, human writers usually start with a draft and keep polishing and revising it. 
As C. J. Cherryh once said, ``it is perfectly okay to write garbage - as long as you edit brilliantly." \footnote{https://www.goodreads.com/quotes/398754-it-is-perfectly-okay-to-write-garbage--as-long-as-you}
We model abstract generation as a conditioned, iterative text generation problem and design a new \textbf{Writing-editing Network} with an \textbf{Attentive Revision Gate} to iteratively examine, improve, and edit the abstract with guidance from the paper title as well as the previously generated abstract. A result of the Writing-editing Network is shown in Table~\ref{table:abstracts}, where we can see that the initial draft contains more topically relevant and richer concepts than the title, such as the term `\textit{IE}'. 
By adding this initial draft as feedback and guidance, it eases the next generation iteration, allowing the model to focus on a more limited learning space, and generate more concise and coherent abstracts. 

\begin{figure*}[!htb]
\centering\small
\includegraphics[width=.9\textwidth]{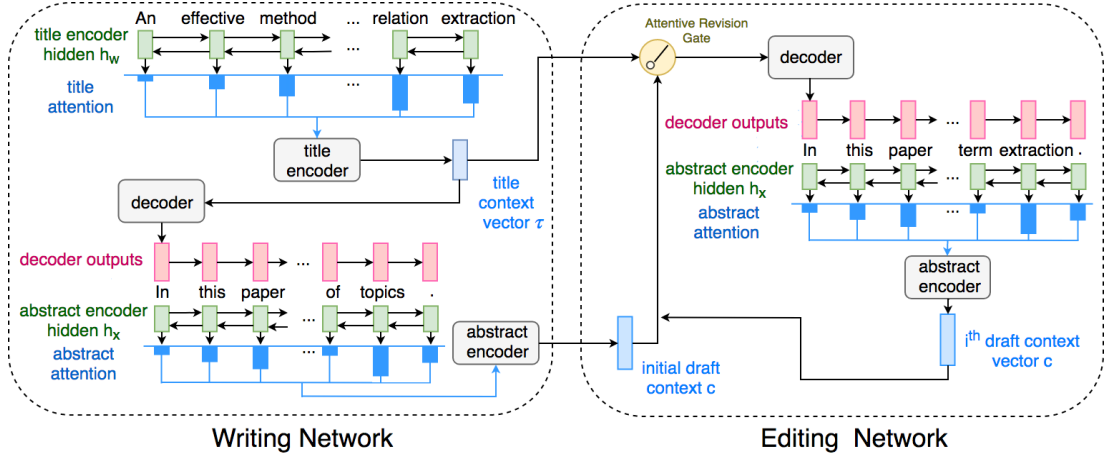}
\caption{Writing-editing Network architecture overview.}
\label{fig:overview}
\end{figure*}
\begin{table*}[!htb]
\footnotesize\small
\begin{tabular}{|p{1.5cm}|p{13.5cm}|}
\hline 
Title & An effective method of using \textbf{Web} based \textbf{information} for \textbf{Relation Extraction~\cite{Keong2008}}\\
\hline 
Human written abstract&\textit{We propose a method that incorporates paraphrase \textbf{information} from the \textbf{Web} to boost the performance of a supervised \textbf{relation extraction} system. Contextual \textbf{information} is extracted from the \textbf{Web} using a semi-supervised process, and summarized by skip-bigram overlap measures over the entire \textbf{extract}. This allows the capture of local contextual \textbf{information} as well as more distant associations. We observe a statistically significant boost in \textbf{relation extraction} performance.}\\
\hline 
LSTM\_LM &\textit{This paper proposes a method for automatic \textbf{extraction} of salient \textbf{information} from an original text. Our method shows promising results in which a noun clustering is employed. The \textbf{relation} is then be applied to the \textbf{relation extraction} task.}\\
\hline
Seq2seq with attention (Initial Draft) &\textit{In this paper we focus on the task of \textbf{extracting} the most commonly used \textbf{Web} pages in named entity recognition (IE) \textbf{information} from a large corpus of news articles. In particular, we propose a new kernel method that can be applied to a large collection of \textbf{Web} pages by automatic detection of labeled and unlabeled data. Our method is very simple and because the related method of using the data to be used as a guide as a source of \textbf{information} for the topic detection and classification can be applied, but often fail at improving the performance by combining \textbf{relations} and previous methods. In this method, the co-occurrence counts and the occurrences of each \textbf{relation} are not statistically significant, but also that it is possible to contribute to entity \textbf{relations} with a term in a variety of topics.}\\
\hline
Writing-editing Networks (Final Draft)&\textit{In this paper we focus on the task of identifying the most commonly relevant features of \textbf{Web} documents. In particular, we propose a generic, automated IE algorithm that can be applied to a large collection of \textbf{Web} pages containing full large documents. This is a first step in helping a wide range of collaborative works for \textbf{relation extraction}. We show that it is possible to eliminate a good number of errors in \textbf{relation extraction} from a variety of documents, but that it is difficult to define a problem of term \textbf{extraction}.}\\
\hline
\end{tabular}
\caption{Human and system generated abstracts for the same title.}
\label{table:abstracts}
\end{table*}

\section{Approach}
In this section, we describe our ``Writing-editing Network'' (Figure~\ref{fig:overview}). The writing network takes a title as input and generates the first abstract draft. The editing network takes both the title and previous draft as input to iteratively proof-read, improve, and generate new versions.
\subsection{Writing Network}
 \label{Writing networks}
 
Our Writing Network is based on an attentive sequence-to-sequence model. We use a bi-directional gated recurrent unit (GRU)~\citep{cho2014learning} as an encoder, which takes a title $\mathcal{T}=\{w_1,\dots,w_K\}$ as input. For each token, $w_k$, the encoder produces a hidden state, $h_{w_k}$. \\
\indent We employ a GRU as our decoder to generate the draft abstract $X^{(0)}=\{x^{(0)}_1,\dots, x^{(0)}_N\}$. To capture the correlation between the title, $\mathcal{T}$, and the abstract draft, $X^{(0)}$, we adopt a soft-alignment attention mechanism \citep{atten14}, which enables the decoder to focus on the most relevant words from the title. At the $n^{\text{th}}$ decoder step, we apply the soft attention to the encoder hidden states to obtain an attentive title context vector, $\tau_{n}$:
\begin{align}
\label{eq:atten}
  \begin{split}
    &\tau_{n} = \sum_{k=1}^{K} \alpha_{n,k}h_{w_k} \\
    &\alpha_{n,k} = \text{softmax}\left(f\left(s_{n-1},h_{w_k}\right)\right)
  \end{split}
\end{align}
where $s_{n-1}$ is the ${n-1}^{th}$ hidden state, $s_0=h_{w_K}$ which is the last hidden state of the encoder, $f$ is a function that measures the relatedness of word $w_k$ in the title and word $x^{(0)}_{n-1}$ in the output abstract. The decoder then generates the $n^{th}$ hidden state, $s_n$, which is given by:
\begin{align}\label{eq:dec}
  \begin{split}
    &s_n=\text{GRU}(x^{(0)}_{n-1},s_{n-1},\tau_n) \\
    &p(x^{(0)}_n|x^{(0)}_{1:n-1},w_{1:K})=g(x^{(0)}_{n-1},s_n, \tau_n)
  \end{split}
\end{align}
where the function $g$ is a softmax classifier, which is used to find the next word, $x^{(0)}_n$, by selecting the word of maximum probability. 

\subsection{Editing Network}
The concepts contained in the titles are usually limited, so the learning space for the generator is huge, which hinders the quality of the generated abstract. Compared to the title, the generated abstracts contain more topically relevant concepts, and can provide better guidance. Therefore, we design an Editing Network, which, besides the title, also takes the previously generated abstract as input and iteratively refines the generated abstract. 
The Editing Network follows an architecture similar to the Writing Network. 

Given an initial draft, $X^{(0)}$, from the Writing Network, 
we use a separate bi-directional GRU encoder, to encode each $x^{(0)}_n\in X^{(0)}$ into a new representation, $h_{x^{(0)}_n}$. As in the Writing Network, we use $s_0=h_{w_K}$ as the initial decoder hidden state of the Editing Network decoder, which shares weights with the Writing Network decoder.\\
\indent At the $n^{\text{th}}$ decoder step, we compute an attentive draft context vector, $c_t$, by applying the same soft attention function from Eq.~\eqref{eq:atten} to the encoded draft representations, $\{h_{x^{(0)}_1},\dots,h_{x^{(0)}_N}\}$, using decoder state $s_{n-1}$.\footnote{The indices are changed since the generated sequence lengths from the writing and editing networks may differ.} We also recompute the attentive title context vector, $\tau_n$, with the same soft attention, though these attentions do not share weights. Intuitively, this attention mechanism allows the model to proofread the previously generated abstract and improve it by better capturing long-term dependency and relevance to the title.
We incorporate $c_t$ into the model through a novel \textbf{Attentive Revision Gate} that adaptively attends to the title and the previous draft at each generation step:
\begin{align}
  \label{eq:gate}
  &r_n = \sigma\left(W_{r,c}c_n + W_{r,\tau}\tau_n + b_r\right)\\
  &z_n = \sigma\left(W_{z,c}c_n + W_{z,\tau}\tau_n + b_z\right)\\
  &\rho_n = \tanh\left(W_{\rho,c}c_n + z_n \odot \left(W_{\rho,\tau}\tau_n + b_\rho\right)\right)\\
  \label{eq:attenc}
  & a_n =   r_n\odot c_n + (1 - r_n)\odot\rho_n
\end{align}
where all $W$ and $b$ are learned parameters. With the attention vector, $a_n$, we compute the $n^{\text{th}}$ token with the same decoder as in section~\ref{Writing networks}, yielding another draft $X^{(1)}=\{x^{(1)}_1,\dots, x^{(1)}_T\}$. We repeat this process for $d$ iterations. In our experiments, we set $d$ to 2 and found it to work best.
\section{Experiments}
\subsection{Data and Hyperparameters}
\label{data}
We select NLP as our test domain because we have easy access to data and domain experts for human judges. We collected a data set of 10,874 paper title and abstract pairs\footnote{\url{https://github.com/EagleW/ACL_titles_abstracts_dataset}} from the ACL Anthology Network\footnote{http://clair.eecs.umich.edu/aan/index.php} (until 2016) for our experiments. We randomly dividing them into training ($80\%$), validation ($10\%$), and testing ($10\%$) sets. On average, each title and abstract include 9 and 116 words, respectively. Our model has 512 dimensional word embeddings, 512 encoder hidden units, and 1,024 decoder hidden units.
\subsection{Method Comparison}

\label{Baselines}

\begin{table}[!htb]
\footnotesize
  \begin{center}
   \begin{tabular}{|p{1.5cm}|p{1.3cm}<{\centering}|p{1.5cm}<{\centering}|p{1.5cm}<{\centering}|}
\hline
\textbf{}       & \textbf{}       & \textbf{}        & \textbf{HUMAN}   \\
\textbf{Method} & \textbf{METEOR} & \textbf{ROUGE-L} & \textbf{PREFER-} \\
\textbf{}       & \textbf{}       & \textbf{}        & \textbf{ENCE}    \\ \hline
LSTM-LM         & 8.7             & 15.1             & 0                             \\ \hline
Seq2seq         & 13.5            & 19.2             & 22                            \\ \hline
ED(1)   		& 13.3   		  & \textbf{20.3}    & 30                            \\ \hline
ED(2)   		& \textbf{14.0}   & 19.8    		 & \textbf{48}                   \\ \hline
\end{tabular}
  \end{center}
\caption{ Method Comparison (\%).}
\label{table:methodcomparison}
\end{table}
\begin{table}[!htb]
\footnotesize
\centering
    \begin{tabular}{|c|c|c|c|c|c|c|}
    \hline
     $n$    & \bf 1  & \bf 2  &\bf 3 &\bf 4  &\bf 5  & \bf 6 \\ \hline
    \bf System & $100$ & $94.4$ & $67.3$ & $35.0$ & $15.9$ & $6.6$ \\ \hline
    \bf Human  & $98.2$ & $78.5$ & $42.2$ & $17.9$ & $7.7$  & $4.1$ \\ \hline
\end{tabular}
\caption{Plagiarism Check: Percentage (\%) of $n$-grams in test abstracts generated by system/human which appeared in training data.}
\label{plag_check}
\end{table}
We include an LSTM Language Model~\citep{lstmlm12} (\textit{LSTM-LM}) and a Seq2seq with Attention (\textit{Seq2seq}) model as our baselines and compare them with the first (\textit{ED(1)}) and second revised draft (\textit{ED(2)}) produced by the Writing-editing Network.

\begin{table*}[!htbp]
\centering
\footnotesize
\begin{tabular}{|p{2.3cm}<{\centering}|p{1.2cm}<{\centering}|p{1.6cm}<{\centering}|p{1.6cm}<{\centering}|p{1.3cm}<{\centering}|p{1.6cm}<{\centering}|p{1.4cm}<{\centering}|p{1cm}<{\centering}|}
\hline
                                  & \textbf{\# Tests} & \textbf{\# Choices} & \multicolumn{2}{c|}{\textbf{Non-expert}}         & \multicolumn{2}{c|}{\textbf{NLP Expert}}            \\ \cline{4-7} 
                                  &                   & \textbf{per Test}   & \textbf{\textbf{Non-CS}} & \textbf{\textbf{CS}} & \textbf{\textbf{Junior}} & \textbf{\textbf{Senior}} \\ \hline
\multirow{3}{*}{Different Titles} & 50                & 2                   & 30\%                     & 15\%                 & 12\%                     & 0\%                      \\ \cline{2-7} 
                                  & 20                & 5                   & 60\%                     & 20\%                 & 30\%                     & 20\%                     \\ \cline{2-7} 
                                  & 10                & 10                  & 80\%                     & 30\%                 & 30\%                     & 20\%                     \\ \hline
\multirow{2}{*}{Same Title}       & 50                & 2                   & 54\%                     & 10\%                 & 4\%                      & 0\%                      \\ \cline{2-7} 
                                  & 20                & 5                   & 75\%                     & 25\%                 & 5\%                      & 5\%                      \\ \hline
\end{tabular}
\caption{\label{table:hresults} Turing Test Passing Rates.}
\end{table*}

Table~\ref{table:methodcomparison} presents METEOR~\citep{denkowski2014meteor} and ROUGE-L~\citep{lin2004rouge} scores for each method, where we can see score gains on both metrics from the Editing Mechanism.  Additionally, 10 NLP researchers manually assess the quality of each method. We randomly selected 50 titles and applied each model to generate an abstract. We then asked human judges to choose the best generated abstract for each title and computed the overall percentage of each model being preferred by human, which we record as \emph{Human Preference}. The criteria the human judges adopt include topical relevance, logical coherence, and conciseness. Table~\ref{table:methodcomparison} shows that the human judges strongly favor the abstracts from our ED(2) method. 

We also conduct a plagiarism check in Table~\ref{plag_check}, which shows that 93.4\% of 6-grams generated by  ED(2) did not appear in the training data, indicating that our model is not simply copying. The 6-grams borrowed by both our model and human include ``\textit{word sense disambiguation ( wsd )}", 
``\textit{support vector machines ( svm )}'', 
``\textit{show that our approach is feasible}'', and ``\textit{we present a machine learning approach}''. 
However, human writing is still more creative. 
The uni-grams and bi-grams that appear in human written test abstracts but not in the training set include ``\textit{android}'', ``\textit{ascii}'', 
`\textit{p2p}'', ``\textit{embellish}'', 
``\textit{supervision bottleneck}'', ``\textit{medical image}'', ``\textit{online behaviors}'', and ``\textit{1,000 languages}''.

\subsection{Impact of Editing Mechanism}
\begin{table}[!htb]
\small
\centering
    \begin{tabular}{|c|c|c|c|c|c|c|}
    \hline
     $n$    	& \bf 1     & \bf 2     &\bf 3   & \bf 4  &\bf 5  & \bf 6 \\ \hline
    METEOR  & $13.3$    & \bf14.0   & $13.6$ & $13.9$ & $13.8$ & $13.5$ \\ \hline
    ROUGE-L & \bf20.3   & $19.8$ 	 & $18.6$ & $19.2$ & $18.9$ & $18.8$ \\ \hline
\end{tabular}
\caption{Iteration comparison (\%)}
\label{iteration}
\end{table}
We trained and evaluated our editing approach with 1-6 iterations and the experimental results (Table~\ref{iteration}) showed that the second iteration produced the best results. The reason may be as follows. The attentive revision gate incorporates the knowledge from the paper title and the previous generated abstract. As the editing process iterates, the knowledge pool will diverge since in each iteration the generated abstract may introduce some irrelevant information. Empirically the second iteration achieved a good trade-off between good quality of generated abstract and relevance with topics in the title. 

\subsection{Turing Test}
\label{Turing Tests}
We carried out two series of Turing tests, where the human judges were asked to distinguish the fake (system-generated) abstracts from the real (human-written) ones. (1)\textbf{Abstracts for different titles}. We asked the human judges to identify the fake abstract from a set of $N-1$ real ones (i.e., N choose 1 question). A test is passed when a human judge mistakenly chooses a real abstract.
(2) \textbf{Abstracts for the same title}. We asked the human judges to choose the real abstract from a set of $N-1$ fake ones. A test is passed when a human judge mistakenly chooses a fake abstract.
As expected, Table~\ref{table:hresults} shows that people with less domain knowledge are more easily deceived. 
Specifically, non-CS human judges fail at more than half of the 1-to-1 sets for the same titles, which suggests that most of our system generated abstracts follow correct grammar and consistent writing style. Domain experts fail on 1 or 2 sets, mostly because the human written abstracts in those sets don't seem very topically relevant. Additionally, the more abstracts that we provided to human judges, the easier it is to conceal the system generated abstract amongst human generated ones. 
A human is still more intelligent than the machine on this task from many reasons: (1) Machines lack knowledge of the deep connections among scientific knowledge elements and thus produce some fluent but scientifically incorrect concepts like ``\textit{...a translation system to generate a parallel corpus...}'' and  ``\textit{...automatic generation of English verbs...}''. (2) Humans know better about what terms are more important than others in a title. For example, if a language name appears in the title, it must appear in the abstract. We have an automatic term labeling approach, but, unfortunately, its performance (75\% F-score) is not good enough to help the abstract generation. (3) Human written abstracts are generally more specific, concise, and engaging, often containing specific lab names, author names (e.g., ``\textit{Collins proposed...}''), system abbreviations, and terminologies (e.g., ``\textit{Italian complex nominals (cns) of the type n+p+n}''). In contrast, our system occasionally generates too general descriptions like ``\textit{Topic modeling is a research topic in Natural Language Processing.}'' (4) Machines lack common sense knowledge, so a system generated abstract may mention three areas/steps, but only outline two of them. (5) Machines lack logical coherence. A system generated abstract may contain ``\textit{The two languages...}'' and not state which languages. (6) We are not asking the system to perform scientific experiments, and thus the system generated ``experimental results'' are often invalid, such as ``\textit{Our system ranked first out of the participating teams in the field of providing such a distribution.}''.

\section{Related work}



Deep neural networks are widely applied to text generation tasks such as poetry creation~\citep{kk1, kk2, creativepoetry17}, recipe generation~\citep{checklist16}, abstractive summarization~\citep{copynet16,absg16,hybridp17}, and biography generation~\citep{biogen16,table2text17}. We introduce a new task of generating paper abstracts from the given titles. We design a Writing-editing Network which shares ideas with Curriculum Learning~\citep{bengio2009curriculum}, where training on a data point from coarse to fine-grained can lead to better convergence~\citep{krueger2009flexible}. 
Our model is different from previous theme-rewriting~\citep{rewriting15,rewriting16} approach which has been applied to math word problems but more similar to the Feedback Network~\citep{16feedback} by using previous generated outputs as feedback to guide subsequent generation. Moreover, our Writing-editing Network treats previous drafts as independent observations and does not propagate errors to previous draft generation stages. This property is vital for training feedback architectures for discrete data. Another similar approach is the deliberation network used for Machine Translation~\citep{deliberation}. 
Instead of directly concatenating the output of the encoder and writing network, we use the learnable Attentive Revision Gate to control their integration.
\section{Conclusions and Future Work}

We propose a new paper abstract generation task, present a novel Writing-editing Network architecture based on an Editing Mechanism, and demonstrate its effectiveness through both automatic and human evaluations. In the future we plan to extend the scope to generate a full paper by taking additional knowledge bases as input.
\bibliography{acl2018}
\bibliographystyle{acl_natbib}
\appendix

\end{document}